\RequirePackage{amsmath}
\documentclass[runningheads,a4paper]{llncs}
\usepackage[T1]{fontenc}
\usepackage{graphicx}
\usepackage{tabularx}
\usepackage{amsmath}
\usepackage[]{hyperref}
\usepackage{booktabs}

\usepackage{siunitx}

\DeclareMathOperator*{\argmax}{arg\,max}

\begin{document}
\title{Towards Automated Customer Support}
\titlerunning{Towards Automated Customer Support}
\author{Momchil Hardalov\inst{1} \and
Ivan Koychev\inst{1} \and
Preslav Nakov\inst{2}}
%
%  et al.
\authorrunning{M. Hardalov, et al.}
\institute{FMI, Sofia University ,,St. Kliment Ohridski'', Sofia, Bulgaria,\\
\email{\{hardalov,koychev\}@fmi.uni-sofia.bg}
\vspace{2pt}
\and
Qatar Computing Research Institute, HBKU, Doha, Qatar,\\
\email{pnakov@qf.org.qa}}
\maketitle              % typeset the header of the contribution
\begin{abstract}

Recent years have seen growing interest in conversational agents, such as chatbots, which are a very good fit for automated customer support because the domain in which they need to operate is narrow. This interest was in part inspired by recent advances in neural machine translation, esp. the rise of sequence-to-sequence (seq2seq) and attention-based models such as the Transformer, which have been applied to various other tasks and have opened new research directions in question answering, chatbots, and conversational systems. Still, in many cases, it might be feasible and even preferable to use simple information retrieval techniques. Thus, here we compare three different models: (\emph{i})~a retrieval model, (\emph{ii})~a sequence-to-sequence model with attention, and (\emph{iii})~Transformer. Our experiments with the Twitter Customer Support Dataset, which contains over two million posts from customer support services of twenty major brands, show that the seq2seq model outperforms the other two in terms of semantics and word overlap.

%In this paper we compare information retrieval and neural network-based approaches for automating the customer support task using the Twitter Customer Support Dataset, a dataset containing over 2 million posts, coming from over 20 big brands. We compare three different models - BM25 for IR model, and two network architectures (Attentive Sequence-to-Sequence and Transformer). Experiments show that Seq2Seq model outperforms convincingly the others in both semantic aspect and word-overlap. Our work takes advantage of metrics for automatic evaluation and does not need human interference.

% The abstract should summarize the contents of the paper
% using at least 70 and at most 150 words. It will be set in 9-point
% font size and be inset 1.0 cm from the right and left margins.
% There will be two blank lines before and after the Abstract. \dots
%
%
%

\end{abstract}

\section{Introduction}

The rapid proliferation of mobile and portable devices has enabled a number of new products and services. Yet, it has also laid stress on customer support as users now also expect 24x7 availability of information about their orders, or answers to basic questions such as `Why is my Internet connection dead?' and `What time is the next train from Sofia to Varna?'

Customer support has always been important to companies. Traditionally, it was offered primarily over the phone, but recently a number of alternative communication channels have emerged such as e-mail, social networks, forums/message boards, live chat, self-serve knowledge base, etc. As a result, it has become increasingly expensive for companies to maintain quality customer support services over a growing number of channels. First, they must find people that have both good language and communication skills. Second, each new employee must go through several training sessions before being able to operate in the target channel, which is inefficient and time-consuming. And finally, it is difficult to have employees available for customer support 24x7.

\noindent Chatbots are especially fit for the task as they are automatic: fully or partially. Moreover, from a technological viewpoint, they are feasible as the domain they need to operate in is narrow. As a result, chit-chat is reduced to a minimum, and chatbots serve primarily as question-answering devices. Moreover, it is possible to train them on real-world chat logs. Here, we experiment with such logs from customer support on Twitter, and we compare two types of chatbots: (\emph{i})~based on information retrieval (IR), and (\emph{ii})~on neural question answering. 
We further explore semantic similarity measures since generic ones such as ROUGE \cite{lin2004rouge}, BLEU \cite{papineni2002bleu} and METEOR \cite{banerjee2005meteor}, which come from machine translation or text summarization, are not well suited for chatbots.

The remainder of this paper is organized as follows: Section~\ref{sec:related} presents related work. Section~\ref{sec:data} describes the dataset and the preprocessing, and offers insights about the dialogs. Section~\ref{sec:methods} focuses on the models. Section~\ref{sec:Experiments} describes the experiments, the results, and the evaluation measures. Section~\ref{sec:discuss} discusses the results. Finally, Section~\ref{sec:conclusion} concludes, and suggests directions for future work.

\section{Related Word}
\label{sec:related}

Sequence-to-sequence (seq2seq) models were first introduced in 2014 in the context of machine translation \cite{sutskever2014sequence}. Since then, they have been successfully applied in other domains such as text summarization, question-answering, conversational agents, etc. One of the first examples of a basic seq2seq model for chatbots was proposed in 2015 by Vinyals et al.~\cite{DBLP:journals/corr/VinyalsL15}, who experimented with two datasets: IT helpdesk tickets and Open Subtitles. They further pointed out to the following issues: lack of context modeling for multi-turn dialogs, lack of ``personality'' for models trained on different sources, and need for human evaluation of the generated responses. 

Another source of training data have been community Question Answering forums. In 2015, Lowe at el. \cite{lowe2015ubuntu} introduced the Ubuntu Dialog Corpus, and experimented with plain RNN vs. LSTM-based neural models, in addition to retrieval-based approaches. An extension of this study, including several new encoder-decoder architectures, was published recently \cite{lowe2017training}. In another related work, Boyanov et al. \cite{boyanov-EtAl:2017:RANLP} explored the utility of neural models on data from SemEval-2016 task 3 on Community Question Answering \cite{nakov-EtAl:2016:SemEval}. They compared seq2seq models with retrieval-based ones, performing model selection using question answering measures, and studied the ability of the chatbot to answer free-form questions.

Twitter data is particularly suitable for fitting neural conversational models because of the length restriction, which encourages people to write short, more precise tweets. Thus, it was used in a number of studies. Serban et al. \cite{serban2015hierarchical} improved seq2seq models using a hierarchical structure. Sordoni et al. \cite{conf/naacl/SordoniGABJMNGD15} worked on modeling the context. Shang et al. \cite{shang2015neural} proposed a neural network response generator for short-text conversation, which was trained with a large number of one-round conversations from a micro-blogging service, and could generate grammatically correct and content-wise appropriate responses. 

\noindent Some interesting approaches for building customer support chatbots were shown in \cite{cui2017superagent,qiu2017alime}, as a combination of retrieval and neural models. Cui et al. \cite{cui2017superagent} used information from in-page product descriptions, as well as user-generated content from e-commerce web sites to improve online shopping experience. Their approach incorporated four different components (fact database, FAQs, opinion-oriented answers, and a neural-based chit-chat generator) into a meta-engine that makes a choice between them. Qiu et al. \cite{qiu2017alime} proposed an open-domain chatbot engine that integrates results from IR and seq2seq models, and uses an attentive seq2seq reranker to choose dynamically between their outputs.

% Twitter \cite{conf/naacl/SordoniGABJMNGD15} \cite{shang2015neural} \\
% Hybrid approaches \cite{qiu2017alime} \cite{cui2017superagent} \\
% Customer Support \cite{di2004bootstrapping} \\
% Ubuntu Dialog \cite{lowe2015ubuntu} \cite{lowe2017training} \\
% Movies \cite{serban2015hierarchical} \\
% Other \cite{serban2017hierarchical} \cite{DBLP:journals/corr/VinyalsL15} \\
% Forum Data QA \cite{boyanov-EtAl:2017:RANLP} \\
% Context \cite{conf/naacl/SordoniGABJMNGD15} \cite{Sordoni:2015:HRE:2806416.2806493}

\section{Dataset}
\label{sec:data}

Overall, data and resources that could be used to train a customer support chatbot are very scarce, as companies keep conversations locked on their own closet, proprietary support systems. This is due to customer privacy concerns and to companies not wanting to make public their know-how and the common issues about their products and services. An extensive 2015 survey on available dialog corpora by Serban et al.~\cite{serban2018survey} found no good publicly available dataset for real-world customer support. 

\begin{figure}[tbh]
\includegraphics[keepaspectratio,width=\textwidth]{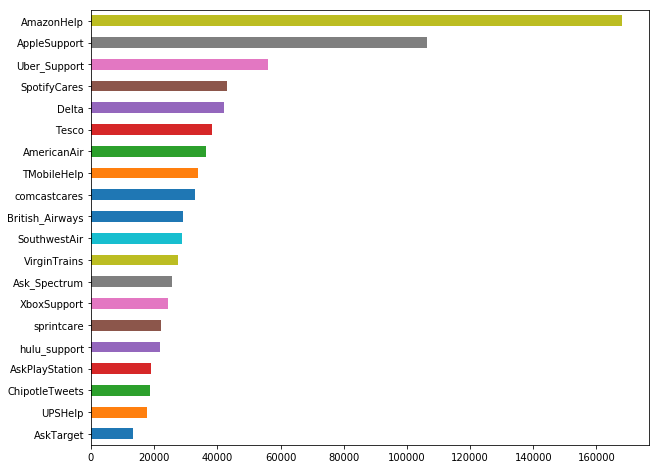}
\caption{Number of user tweets with replies from customer support per company.}
\label{fig:authors}
\end{figure}

\noindent Earlier this year, this situation has changed as a new open dataset, named \emph{Customer Support on Twitter}, was made available on Kaggle.\footnote{\url{https://www.kaggle.com/thoughtvector/customer-support-on-twitter}} It is a large corpus of recent tweets and replies, which is designed to support innovation in natural language understanding and conversational models, and to help study modern customer support practices and impact.
The dataset contains 3M tweets from 20 big companies such as Amazon, Apple, Uber, Delta, and Spotify, among others. See Figure~\ref{fig:authors} for detail.

As customer support topics from different organizations are generally unrelated to each other, we focus only on tweets related to Apple support, which represents the largest number of tweets in the corpus. 
%This allows us to stay focused on a small range of topics that are related to a single company, a situation closer to a real-world scenario. 
We filtered all utterances that redirect the user to another communication channel, e.g.,~direct messages, which are not informative for the model and only bring noise. Moreover, since answers evolve over time, we divided our dataset into a training and a testing part, keeping earlier posts for training and the latest ones for testing. We further excluded from the training set all conversations that are older then sixty days. For evaluation, we used dialogs from the last five days in the dataset, to simulate a real-world scenario for customer support. We ended up with a dataset of 49,626 dialog tuples divided in 45,582 for training and 4,044 for testing. 

Table~\ref{tab:dialogstats} shows some statistics about our dataset. In the top of the table, we can see that the average number of turns per dialog is under three, which means that most of the dialogs finish after one answer from customer support. The bottom of the table shows the distribution of words in the user questions vs. the customer support answers. We can see that answers tend to be slightly longer, which is natural as replies by customer support must be extensive and helpful.

\begin{table}[h!]
  \centering
  \begin{tabular}{>{\centering\arraybackslash}p{7cm} >{\centering\arraybackslash}p{3cm}}
  \toprule
  \multicolumn{2}{c}{\bf Overall}\\
    \midrule
    \# dialogs tuples & 49,626 \\
    \# words (in total) &  26,140 \\
    Min \# turns per dialog &  $\num[round-mode = places]{2}$ \\
    Max \# turns per dialog &  $\num[round-mode = places]{106}$ \\
    Avg. \# turns per dialog & 2.6 \\
%     Avg. \# words in question &  $\num[round-mode = places]{20.000000}$ \\
%     Avg. \# words in answer &  $\num[round-mode = places]{25.884436}$ \\
	\midrule
    Training set: \# of dialogs & 45,582 \\
    Testing set: \# of dialogs & 4,044 \\
    %\bottomrule
\end{tabular}

\begin{tabular}{>{\centering\arraybackslash}p{4cm} >{\centering\arraybackslash}p{3cm} >{\centering\arraybackslash}p{3cm}}
	\toprule
	& \bf Questions & \bf Answers \\
    \midrule
    Avg. \# words      & $\num[round-mode = places]{21.314814}$ &    $\num[round-mode = places]{25.884436}$ \\
    %std      &  $\num[round-mode = places]{11.353460}$ &  $\num[round-mode = places]{9.523078}$ \\
    Min \# words      &  $\num[round-mode = places]{1.000000}$ &    $\num[round-mode = places]{3.000000}$ \\
    1st quantile   &   $\num[round-mode = places]{13.000000}$ &    $\num[round-mode = places]{20.000000}$ \\ 
    Mode  &     $\num[round-mode = places]{20.000000}$ &   $\num[round-mode = places]{23.000000}$ \\
    3rd quantile &    $\num[round-mode = places]{ 27.000000}$ &   $\num[round-mode = places]{29.000000}$ \\
    Max \# words   &     $\num[round-mode = places]{136.000000}$ &  $\num[round-mode = places]{70.000000}$ \\
    \bottomrule
	\end{tabular}
  \caption{Statistics about our dataset.}
  \label{tab:dialogstats}
\end{table}

% Questions
% mean        21.314814
% std         11.353460
% min          1.000000
% 25%         13.000000
% 50%         20.000000
% 75%         27.000000
% max        136.000000

%Answers 
% mean        25.884436
% std          9.523078
% min          3.000000
% 25%         20.000000
% 50%         23.000000
% 75%         29.000000
% max         70.000000

\section{Method}
\label{sec:methods}

\subsection{Preprocessing}
Since Twitter has its own specifics of writing in terms of both length\footnote{By design, tweets have been strictly limited to 140 characters; this constrain has been relaxed to 280 characters in 2017.} and style, standard text tokenization is generally not suitable for tweets. Therefore, we used a specialized Twitter tokenizer \cite{manning2014stanford} to preprocess the data. Then, we further cleaned the data by replacing the shorthand entries, e.g., \emph{'ll}, \emph{'d}, \emph{'re}, \emph{'ve}, with the most likely literary form, e.g., \emph{will}, \emph{would}, \emph{are}, \emph{have}. We also replaced slang words, e.g., \emph{'bout} and \emph{'til}, with the standard words, e.g., \emph{about} and \emph{until}. Similarly, we replaced URLs with the special word \textit{<url>}, all user mentions with \textit{<user>}, and all hashtags with\textit{<hashtag>}.

Moreover, we tried to mitigate the effect of missing context in long conversations by concatenating all previous turns to the current question. Finally, since seq2seq models cannot be trained efficiently with a large vocabulary, we chose the top $N$ words when building the model (see Section~\ref{sec:Experiments} for detail), and we replaced the instances of the remaining words with a special symbol \textit{<unk>}.\footnote{In future work, we plan to try byte-pair encoding instead \cite{sennrich-haddow-birch:2016:P16-12}.}

\subsection{Information Retrieval}

The Information Retrieval (IR) approach can be defined as follows: given a user question $q^{\prime}$ and a list of pairs of previously asked questions and their answers $(Q,A)= \{(q_j,a_j)|j=1,\ldots,n\}$,
find the most similar question $q_i$ in the training dataset that a user has previously asked and return the answer $a_i$ that customer support has given to $q_i$. The similarity between $q^{\prime}$ and $q_i$ can be calculated in various ways, but most commonly this is done using the cosine between the corresponding TF.IDF-weighted vectors.

\begin{equation}
a^{\prime} = \argmax_{(q_j,a_j)} sim(q^{\prime}, q_j) 
\label{eq:bestrank}
\end{equation}

%the document-based chatbot retrieves a answer $\mathcal{a}$, by finding the most similar question in the corpus (Twitter posts) $\mathcal{Q}$. One common similarity metric used in practice is the measure of cosine angle between the two query vectors. After obtaining ranikings for each $\mathcal{q}$ in the question set, the conversational agent  top ranked (based on its similarity score) $\mathcal{\hat{a}}$ is selected:

% question $\mathcal{q\prime}$ and a response set (Twitter posts) $\mathcal{A}$, the document-based chatbot retrieves a suitable answer $\mathcal{\hat{a}}$ from an inverted index. First it retrieves response candidates from $\mathcal{A}$ based on $\mathcal{q\prime}$ by evaluating some similarity measure between the questions, e.g., cosine similarity. Each entry $\mathcal{q} \in \mathcal{Q}$ is a previously generated answer by the customer support. Each $\mathcal{A}$ is then ranked, and the best candidate $\mathcal{\hat{A}}$ is selected based on its score:
 
% \begin{equation}
% \mathcal{\hat{a}} = \argmax_{\mathcal{q} \in \mathcal{Q}}{Rank(\mathcal{q}, \mathcal{q^\prime})}
% \label{eq:bestrank}
% \end{equation}

\subsection{Sequence-to-Sequence}

Our encoder uses а bidirectional recurrent neural network (RNN) based on long short-term memory (LSTM) \cite{Hochreiter:1997:LSM:1246443.1246450}. It encodes the input sequence $x = (x_1, \dots, x_n)$ and calculates a forward sequence of hidden states $(\overrightarrow{h_1}, \dots, \overrightarrow{h_m})$ and also a backward sequence $(\overleftarrow{h_1}, \dots, \overleftarrow{h_m})$. The decoder is a unidirectional LSTM-based RNN, and it predicts the output sequence $y = (y_1, \dots, y_n)$. Each $y_i$ is predicted using the recurrent state $s_i$, the previous predicted word $y_{i-1}$, and a context vector $c_i$. The latter is computed using an attention mechanism as a weighted sum over the encoder's output $(\overrightarrow{h_j},\overleftarrow{h_j})$, as proposed by Bahdanau et. al \cite{bahdanau2014neural}.

\subsection{Transformer}
The Transformer model was proposed in 2017 by Vaswani et al. \cite{NIPS2017_7181}, and it has shown very strong performance for machine translation, e.g., it achieved state-of-the-art results on WMT2014 data for English-German and English-French translation. Similarly to the seqseq model, the Transformer has an encoder and a decoder. The encoder is a stack of identical layers, based on multi-head self-attention and a simple position-wise fully connected network. The decoder is similar, but in addition to the two sub-layers in the encoder, it introduces a third sub-layer, which performs multi-head attention over the encoders' stack outputs. The main advantage of the Transformer model is that it can be trained significantly faster, as compared to recurrent or convolutional networks. 

\section{Experiments}
\label{sec:Experiments}
%freq = 4
% seq 1 words: 7939
% seq 2 words: 2589
% seq 1 words: 7943
% seq 2 words: 2593
% d_model = 256
% s2s = Transformer(itokens, otokens, len_limit=70, d_model=d_model, d_inner_hid=512, \
%         n_head=4, d_k=64, d_v=64, layers=2, dropout=0.1)

We performed three experiments using the models described in Section~\ref{sec:methods}. Below, each model is abbreviated by its architecture name from \ref{subsec:results}.

\textit{IR} is based on ElasticSearch\footnote{\url{https://www.elastic.co/products/elasticsearch}} (ES), as it provides out-of-the-box implementation of all the components we need. We fed the pre-processed training data into an index with English analyzer enabled, whitespace- and punctuation-based tokenization, and word 3-grams. For retrieval, we used the default BM25 algorithm~\cite{Robertson:2009:PRF:1704809.1704810}, which is an improved version of TF.IDF. For all training questions and for all testing queries, we appended the previous turns in the dialog as context. Given a user question from the testing set,  we returned the customer support answer for the top-ranked result from ES.

\textit{Seq2Seq} contains one bi-directional LSTM layer with 512 hidden units per direction (a total of 1,024). The decoder has two unidirectional layers connected directly to the bidirectional one in the encoder. The network takes as input words encoded as 200-dimensional embeddings. It is a combination of pre-trained GloVe \cite{pennington2014glove} vectors learned from 27B Twitter posts\footnote{\url{https://nlp.stanford.edu/projects/glove/}} for the known words, and a positional embedding layer, learned as model parameters, for the unknown words. The embedding layers for the encoder and for the decoder are not shared, and are learned separately. This separation is due to the fact that the words used in utterances by customers are very different compared to posts by the support. In our experiments, we used the top 8,192 words sorted by frequency for both the embedding and the output. Based on the statistics presented in Section~\ref{sec:data}, we chose to use 60 words (time-steps) for both the encoder and the decoder. We avoid overfitting by applying dropout \cite{srivastava2014dropout} with keep probability of 0.8 after each recurrent layer. For the optimizer, we used Adam \cite{kingma2015adam} with a base value of $\num{1e-03}$ and an exponential decay of 0.99 per epoch.

\textit{Transformer} is based on two identical layers for the encoder and for the decoder, with four heads for the self-attention. The dimensionality of the input and of the output is $d_{model} = 256$, and the inner dimensionality is $d_{inner} = 512$. The input consists of queries with keys of dimension $d_k = 64$ and values of dimension $d_v = 64$. The input and the output embedding are learned separately with sinusoidal positional encoding. The dropout is set to 0.9 keep probability. For the optimization, we use Adam with varying learning rate based on eq. (\ref{eq:aiaynlr}). The hyper-parameter choice was guided by the experiments described by the authors in the original Transformer paper \cite{NIPS2017_7181}. 
\begin{equation}
lrate = d_{model}^{-0.5} \cdot \min{(step\_num^{-0.5}, step\_num \cdot warmup\_steps^{-1.5})}
\label{eq:aiaynlr}
\end{equation}

\subsection{Evaluation Measures}

%The task of dialogue generation is different than the machine translation one. Customer support chatbots do nоt have to generate an exact response, but a similar one with the same semantic meaning. 

How to evaluate a chatbot is an open research question. As the problem is related to machine translation (MT) and text summarization (TS), which are nowadays also addressed using seq2seq models, researchers have been using MT and TS evaluation measures such as BLEU~\cite{papineni2002bleu}, ROUGE~\cite{lin2004rouge}, and METEOR~\cite{banerjee2005meteor}, which focus primarily on word overlap and measure the similarity between the chatbot's response and the gold customer support answer to the user question.
However, it has been argued \cite{liu-EtAl:2016:EMNLP20163,lowe-EtAl:2017:Long} that such word-overlapping measures are not very suitable for evaluating chatbots. Thus, we adopt three additional measures, which are more semantic in nature.\footnote{Note that we do not use measures trained on the same data as advised by \cite{liu-EtAl:2016:EMNLP20163}.} 

%The semantic similarity metrics aim to construct a descriptive vector for the original utterance, and proposed response from the model. We use three different approaches to gather the similarity vectors.

The \textit{embedding average} constructs a vector for a piece of text by taking the average of the word embeddings of its constituent words. Then, the vectors for the chatbot response and for the gold human one are compared using cosine similarity.

The \textit{greedy matching} was introduced in the context of intelligent tutoring systems \cite{rus2012comparison}. It matches each word in the chatbot output to the most similar word in the gold human response, where the similarity is measured as the cosine between the corresponding word embeddings, multiplied by a weighting term, which we set to 1, as shown in equation (\ref{eq:matching}). Since this measure is asymmetric, we calculate it a second time, with arguments swapped, and then we take the average as shown in equation~\ref{eq:matching_simetric}.

\begin{equation}
\label{eq:matching}
greedy(u_1, u_2) = \frac{
	\sum_{v \in u_1}{} weight(v) * \max_{w \in u_2} cos(v,w)}
    {
    \sum_{v \in u_1}{} weight(v)}
\end{equation}

\begin{equation}
\label{eq:matching_simetric}
simGreedy(u_1, u_2) = \frac{greedy(u_1, u_2) + greedy(u_2, u_1)}{2}
\end{equation}

The \emph{vector extrema} was proposed by Forgues et al. \cite{forgues2014bootstrapping} for dialogue systems. Instead of averaging the word embeddings of the words in a piece of text, it takes the coordinate-wise maximum (or minimum), as shown in equation(\ref{eq:exterma}). Finally, the resulting vectors for the chatbot output and for the gold human one are compared using cosine.

\begin{equation}
\label{eq:exterma}
extrema(u_i) =  
\begin{cases}
    \max u_i, & if \max u_i \geq |\min u_i| \\
    \min u_i, & \text{otherwise}
\end{cases}
\end{equation}

%Seq2Seq + 512 + 2 layer + uni + BahdanauAttention + 2 \^ 14 dict + 0.99 decay + lstm + 0.2 Dropout

\subsection{Results}
\label{subsec:results}

Table \ref{tab:wordoverlap} shows the results for the three models we compare (IR, seq2seq, and Transformer) when using word overlap measures such as BLEU@2, which uses unigrams and bigrams only, and ROUGE-L \cite{lin-och:2004:ACL}, which uses Longest Common Subsequence (LCS).

\begin{table}[!htbp]
  \centering
  \begin{tabular}{>{\centering\arraybackslash}p{3cm} >{\centering\arraybackslash}p{3cm} >{\centering\arraybackslash}p{3cm}}
    \toprule
    & \multicolumn{2}{c}{\textbf{Word Overlap Measures}} \\
    & BLEU@2 & ROUGE-L \\
    \midrule
    IR - BM25 &
    $\num[round-mode = places]{13.732301176466532}$ & 
    $\num[round-mode = places]{22.347860605005877}$ \\
    Seq2Seq & 
    \bf{$\num[round-mode = places]{15.104511595780986}$} & 
    \bf{$\num[round-mode = places]{26.597981802417063}$} \\
    Transformer & 
    $\num[round-mode = places]{12.427277110577412}$ & 
    $\num[round-mode = places]{25.32858959229684}$ \\
    \bottomrule
    \\
  \end{tabular}
  \caption{Results based on word-overlap measures.}
  \label{tab:wordoverlap}
\end{table}

Table \ref{tab:genmetrics} shows the results for the same three systems, but using the above-described semantic evaluation measures, namely Embedding Average (with cosine similarity), Greedy Matching, and Vector Extrema (with cosine similarity). For all three measures, we used Google's pre-trained word2vec embeddings because they are not learned during training, which helps avoid bias, as has been suggested in \cite{liu-EtAl:2016:EMNLP20163,lowe-EtAl:2017:Long}.

\begin{table}[!htbp]
  \centering
  \begin{tabularx}{\textwidth}{p{3cm} p{3cm} p{3cm} p{3cm}}
    \toprule
    & \multicolumn{3}{c}{\textbf{Semantic Evaluation Measures}} \\
    & Embedding Average & Greedy Matching & Vector Extrema \\
    \midrule
    IR - BM25 &  
    $\num[round-mode = places]{76.52658639738192}$ & 
    $\num[round-mode = places]{29.719912622416917}$ &
    $\num[round-mode = places]{37.98667099511575}$ \\
    Seq2Seq & 
    \bf{$\num[round-mode = places]{77.10925516393053}$} &
    \bf{$\num[round-mode = places]{30.809532318667042}$} &
    \bf{$\num[round-mode = places]{40.231592168970955}$} \\
    Transformer &
    $\num[round-mode = places]{75.35035772464175}$ & 
    $\num[round-mode = places]{30.07707395198416}$ &
    $\num[round-mode = places]{39.39895553309098}$ \\
    \bottomrule
    \\
  \end{tabularx}
  \caption{Results based on semantic measures.}
  \label{tab:genmetrics}
\end{table}

\section{Discussion}
\label{sec:discuss}

The evaluation results show that \textit{Seq2Seq} performed best with respect to all five evaluation measures.
For the group of semantic measures, it outperformed the other systems in terms of Embedding Average by $\num[round-mode = places,retain-explicit-plus]{+0.582668767}$, in terms of Greedy Matching by $\num[round-mode = places,retain-explicit-plus]{+0.732458367}$, and in terms of Vector Extrema by $\num[round-mode = places,retain-explicit-plus]{+0.832636636}$ (points absolute). Moreover, SeqSeq was also clearly the best model in terms of word-overlap evaluation measures, scoring $\num[round-mode = places]{15.104511595780986}$ on BLEU@2 ($\num[round-mode = places,retain-explicit-plus]{+1.372210419}$ ahead of the second), and $\num[round-mode = places]{26.597981802417063}$ on ROUGE-L ($\num[round-mode = places,retain-explicit-plus]{+1.26939221}$ compared to the second best system).

\noindent The \textit{Transformer} model was ranked second by three of the evaluation measures: Greedy Matching, Vector Extrema, and ROUGE-L. This was unexpected given the state-of-the-art results it achieved for neural machine translation. Higher Greedy Matching and Vector Extrema scores show that the Transformer was able to capture the semantics of the gold answer. Moreover, lower Embedding Average and BLEU@2 scores suggest that it chose different vocabulary or used different word order. This is confirmed by lower ROUGE-L, which is based on longest common subsequence. 

Finally, the retrieval (\textit{IR}) model achieved the second-best results in terms of BLEU@2 and Embedding Average, but it was the worst according to the other three evaluation measures. This shows the superiority of the generative neural models over simple retrieval.

Table~\ref{tab:convs} shows some example responses generated by the three models. 
In the first example \textit{(1)}, the IR model is off and retrieves an answer that addresses a different customer problem. The Seq2Seq model is on the right track, because it asks the user about his device. The Transformer suggests a similar utterance, but it makes an assumption about the phone's operating system, which was not stated in the user's question. 
In the second example \textit{(2)}, all models propose very different ways of action to the user, compared to the original answer, and they all seem plausible in this context; yet, the Transformer is a bit off. 
The next example \textit{(3)} illustrates the ability of the three models to distinguish between different languages, and point the user in the right direction. 
The last example \textit{(4)} is a typical example when neural models fail. The particular question--answer tuple is hard to answer as there are very few similar examples in the training data. Thus, what the neural models generate ends up being off-topic. In contrast, the retrieval approach was able to overcome this and to propose a very good answer.

\begin{table}[!htbp]
  \begin{tabular} {p{5cm}@{ }@{ }@{ }p{5cm}}

  \toprule
  \bf Question \& Answer & \bf Response \\
  \hline

  \textbf{(1)} my phones speaker just broke down on me thanks <user> do better \newline

  <user> we are here to help. which iphone and ios version are you using? are you having a sound quality issue or is there no sound at all? &
  \textbf{IR:} <user> we can help with your device . do you notice the slowness when performing a particular activity ? \newline
  \textbf{Seq2Seq:} <user> that is not the experience we want you to have . which iphone model do you own ? \newline
  \textbf{Transformer:} <user> thanks for reaching out to us . we are always happy to help . can you tell us which version of ios 11 you are using ? \\
  \hline

  \textbf{(2)} i am trying to type it not fuckin i.t, your product is absolute garbage. figure it out. <user> <user> \newline

  <user> lets take a closer look at this to get your device working right. which model do you have and that is the exact os version number installed?

  have you tried any steps so far and when did this start happening? &
  \textbf{IR:} <user> thank you for reaching out . we want to help . to start , lets go to settings > general > about and see which version of ios your iphone is on please . \newline
  \textbf{Seq2Seq:} <user> we would be happy to look into that is going on . can you tell us more? \newline
  \textbf{Transformer:} <user> heres what you can do to work around the issue until it is fixed in a future software update : <url> \\
  \hline

  \textbf{(3)} capaz que si actualizo ios me tel\'{e}fono se convierta en ladrillo :(....ya no conf\'{i}o en sus productos <user> \newline

  <user> we offer support via twitter in english. get help in spanish here: <url> or join <url> &
  \textbf{All models:} <user> we offer support via twitter in english get help in spanish here <url> or join <url> \\
  \hline

  \textbf{(4)} <user> excuse me , i am got a refund confirmation email but nothing happened . do i have to do anythings ? $\rightarrow$
  <user> thanks for reaching out to us about your refund ! to clarify , was this a refund for a purchase in the itunes store ? or was this a product purchase ? 
  <user> thanks for reply . i am purchased music in itunes store .\newline

  <user> thanks. our itunes store support team would have the best information for you on that. you can contact them here: <url> &
  \textbf{IR:} <user> alright . our itunes support team will be able to look at this with you . please reach out to them here : <url> \newline
  \textbf{Seq2Seq:} <user> we are investigating and will update our system status page as more info becomes available <url> \newline
  \textbf{Transformer:} <user> thanks for reaching out . we would recommend leaving that request on our feedback page : <url> \\
  \bottomrule
  \\
  \end{tabular}
\caption{Chatbot responses. The first column shows the original question and the gold customer support answer, while the second column shows responses by our models.}
\label{tab:convs}
\end{table}

\section{Conclusion and Future Work}
\label{sec:conclusion}

We have presented a study on automating customer support on Twitter using two types of models: (\emph{i})~retrieval-based (IR with BM25), and (\emph{ii})~based on generative neural networks (seq2seq with attention and Transformer). We evaluated these models without the need of human judgments, using evaluation measures based on (\emph{i})~word-overlap (BLEU@2 and ROUGE-L), and (\emph{ii})~semantics (Embedding Average, Greedy Matching, and Vector Extrema).
%For our experiments, we have divided the data by the timestamp of the post in order to simulate a real-world scenario. 
%Our results suggest that Attentive Seq2Seq outperforms other models by 0.6--0.8 points absolute for semantic measures, and over 1.0 point absolute for word overlap measures.
Our experiments have shown that generative neural models outperform retrieval-based ones, but they struggle when very few examples for a particular topic are present in the training data. 

% Despite showing good results and being able to generate grammatically correct answers and mostly relevant to the question answers, the data provided only from chat logs is not enough to build an end-to-end customer support bot. It is due to the evolving nature of customer issues, while being accurate when they were posted, they tend to become obsolete with time. 

In future work, we plan to combine the three approaches into an ensemble. Another interesting direction that we would like to explore is handling questions whose correct answers evolve over time, e.g., due to service updates or to new products being released.

\subsubsection*{Acknowledgments.} This work was supported by the EC under grant no. 763566 and by the Bulgarian National Scientific Fund as project no. DN 12/9,

\bibliographystyle{splncs04}
\bibliography{aimsa_towards_2018}

\begin{thebibliography}{10}
\providecommand{\url}[1]{\texttt{#1}}
\providecommand{\urlprefix}{URL }
\providecommand{\doi}[1]{https://doi.org/#1}

\bibitem{bahdanau2014neural}
Bahdanau, D., Cho, K., Bengio, Y.: Neural machine translation by jointly
  learning to align and translate. arXiv preprint arXiv:1409.0473  (2014)

\bibitem{banerjee2005meteor}
Banerjee, S., Lavie, A.: {METEOR}: An automatic metric for {MT} evaluation with
  improved correlation with human judgments. In: Proceedings of the ACL
  Workshop on Intrinsic and Extrinsic Evaluation Measures for Machine
  Translation and/or Summarization. pp. 65--72. Ann Arbor, Michigan (2005)

\bibitem{boyanov-EtAl:2017:RANLP}
Boyanov, M., Nakov, P., Moschitti, A., Da~San~Martino, G., Koychev, I.:
  Building chatbots from forum data: Model selection using question answering
  metrics. In: Proceedings of the International Conference Recent Advances in
  Natural Language Processing, RANLP 2017. pp. 121--129. Varna, Bulgaria (2017)

\bibitem{cui2017superagent}
Cui, L., Huang, S., Wei, F., Tan, C., Duan, C., Zhou, M.: {SuperAgent}: A
  customer service chatbot for e-commerce websites. In: Proceedings of the
  Association for Computational Linguistics 2017, System Demonstrations. pp.
  97--102. ACL~'17, Vancouver, Canada (2017)

\bibitem{forgues2014bootstrapping}
Forgues, G., Pineau, J., Larchev{\^e}que, J.M., Tremblay, R.: Bootstrapping
  dialog systems with word embeddings. In: Proceedings of the NIPS Workshop on
  Modern Machine Learning and Natural Language Processing. Montreal, Canada
  (2014)

\bibitem{Hochreiter:1997:LSM:1246443.1246450}
Hochreiter, S., Schmidhuber, J.: Long short-term memory. Neural Comput.
  \textbf{9}(8),  1735--1780 (1997)

\bibitem{kingma2015adam}
Kingma, D.P., Ba, J.: Adam: A method for stochastic optimization. In:
  Proceedings of the 2015 International Conference on Learning Representations.
  ICLR~'15, San Diego, California (2015)

\bibitem{lin2004rouge}
Lin, C.Y.: {ROUGE}: A package for automatic evaluation of summaries. In:
  Proceedings of the ACL Workshop on Text Summarization Branches Out. pp.
  74--81. Barcelona, Spain (2004)

\bibitem{lin-och:2004:ACL}
Lin, C.Y., Och, F.J.: Automatic evaluation of machine translation quality using
  longest common subsequence and skip-bigram statistics. In: Proceedings of the
  42nd Annual Conference of the Association for Computational Linguistics. pp.
  605--612. ACL~'04, Barcelona, Spain (2004)

\bibitem{liu-EtAl:2016:EMNLP20163}
Liu, C.W., Lowe, R., Serban, I., Noseworthy, M., Charlin, L., Pineau, J.: How
  {NOT} to evaluate your dialogue system: An empirical study of unsupervised
  evaluation metrics for dialogue response generation. In: Proceedings of the
  2016 Conference on Empirical Methods in Natural Language Processing. pp.
  2122--2132. EMNLP~'16, Austin, Texas (2016)

\bibitem{lowe-EtAl:2017:Long}
Lowe, R., Noseworthy, M., Serban, I.V., Angelard-Gontier, N., Bengio, Y.,
  Pineau, J.: Towards an automatic {T}uring test: Learning to evaluate dialogue
  responses. In: Proceedings of the 55th Annual Meeting of the Association for
  Computational Linguistics. pp. 1116--1126. ACL~'17, Vancouver, Canada (2017)

\bibitem{lowe2015ubuntu}
Lowe, R., Pow, N., Serban, I., Pineau, J.: The {U}buntu dialogue corpus: A
  large dataset for research in unstructured multi-turn dialogue systems. In:
  Proceedings of the 16th Annual Meeting of the Special Interest Group on
  Discourse and Dialogue. pp. 285--294. SIGDIAL~'15, Prague, Czech Republic
  (2015)

\bibitem{lowe2017training}
Lowe, R.T., Pow, N., Serban, I.V., Charlin, L., Liu, C.W., Pineau, J.: Training
  end-to-end dialogue systems with the {U}buntu dialogue corpus. Dialogue \&
  Discourse  \textbf{8}(1),  31--65 (2017)

\bibitem{manning2014stanford}
Manning, C., Surdeanu, M., Bauer, J., Finkel, J., Bethard, S., McClosky, D.:
  The {Stanford CoreNLP} natural language processing toolkit. In: Proceedings
  of 52nd Annual Meeting of the Association for Computational Linguistics:
  System Demonstrations. pp. 55--60. ACL~'14, Baltimore, Maryland (2014)

\bibitem{nakov-EtAl:2016:SemEval}
Nakov, P., M\`{a}rquez, L., Moschitti, A., Magdy, W., Mubarak, H., Freihat,
  a.A., Glass, J., Randeree, B.: {SemEval}-2016 task 3: Community question
  answering. In: Proceedings of the 10th International Workshop on Semantic
  Evaluation. pp. 525--545. SemEval~'16, San Diego, California (2016)

\bibitem{papineni2002bleu}
Papineni, K., Roukos, S., Ward, T., Zhu, W.J.: {BLEU}: A method for automatic
  evaluation of machine translation. In: Proceedings of the 40th Annual Meeting
  of the Association for Computational Linguistics. pp. 311--318. ACL '02,
  Philadelphia, Pennsylvania (2002)

\bibitem{pennington2014glove}
Pennington, J., Socher, R., Manning, C.: {GloVe}: Global vectors for word
  representation. In: Proceedings of the 2014 Conference on Empirical Methods
  in Natural Language Processing. pp. 1532--1543. EMNLP~'14, Doha, Qatar (2014)

\bibitem{qiu2017alime}
Qiu, M., Li, F.L., Wang, S., Gao, X., Chen, Y., Zhao, W., Chen, H., Huang, J.,
  Chu, W.: {AliMe} chat: A sequence to sequence and rerank based chatbot
  engine. In: Proceedings of the 55th Annual Meeting of the Association for
  Computational Linguistics. pp. 498--503. ACL~'17, Vancouver, Canada (2017)

\bibitem{Robertson:2009:PRF:1704809.1704810}
Robertson, S., Zaragoza, H.: The probabilistic relevance framework: {BM25} and
  beyond. Found. Trends Inf. Retr.  \textbf{3}(4),  333--389 (Apr 2009)

\bibitem{rus2012comparison}
Rus, V., Lintean, M.: A comparison of greedy and optimal assessment of natural
  language student input using word-to-word similarity metrics. In: Proceedings
  of the Seventh Workshop on Building Educational Applications Using NLP. pp.
  157--162. Montreal, Canada (2012)

\bibitem{sennrich-haddow-birch:2016:P16-12}
Sennrich, R., Haddow, B., Birch, A.: Neural machine translation of rare words
  with subword units. In: Proceedings of the 54th Annual Meeting of the
  Association for Computational Linguistics. pp. 1715--1725. ACL~'16 (2016)

\bibitem{serban2018survey}
Serban, I.V., Lowe, R., Henderson, P., Charlin, L., Pineau, J.: A survey of
  available corpora for building data-driven dialogue systems: The journal
  version. Dialogue \& Discourse  \textbf{9}(1),  1--49 (2018)

\bibitem{serban2015hierarchical}
Serban, I.V., Sordoni, A., Bengio, Y., Courville, A.C., Pineau, J.:
  Hierarchical neural network generative models for movie dialogues. CoRR,
  abs/1507.04808  (2015)

\bibitem{shang2015neural}
Shang, L., Lu, Z., Li, H.: Neural responding machine for short-text
  conversation. In: Proceedings of the 53rd Annual Meeting of the Association
  for Computational Linguistics and the 7th International Joint Conference on
  Natural Language Processing. pp. 1577--1586. ACL-IJCNLP'15, Beijing, China
  (2015)

\bibitem{conf/naacl/SordoniGABJMNGD15}
Sordoni, A., Galley, M., Auli, M., Brockett, C., Ji, Y., Mitchell, M., Nie,
  J.Y., Gao, J., Dolan, B.: A neural network approach to context-sensitive
  generation of conversational responses. In: Proceedings of the 2015
  Conference of the North American Chapter of the Association for Computational
  Linguistics: Human Language Technologies. pp. 196--205. NAACL-HLT~'15,
  Denver, Colorado (2015)

\bibitem{srivastava2014dropout}
Srivastava, N., Hinton, G., Krizhevsky, A., Sutskever, I., Salakhutdinov, R.:
  Dropout: A simple way to prevent neural networks from overfitting. The
  Journal of Machine Learning Research  \textbf{15}(1),  1929--1958 (2014)

\bibitem{sutskever2014sequence}
Sutskever, I., Vinyals, O., Le, Q.V.: Sequence to sequence learning with neural
  networks. In: Proceedings of the 27th Annual Conference on Neural Information
  Processing Systems. pp. 3104--3112. NIPS~'14, Montreal, Canada (2014)

\bibitem{NIPS2017_7181}
Vaswani, A., Shazeer, N., Parmar, N., Uszkoreit, J., Jones, L., Gomez, A.N.,
  Kaiser, L.u., Polosukhin, I.: Attention is all you need. In: Proceedings of
  the 30th Annual Conference on Neural Information Processing Systems. pp.
  5998--6008. NIPS~'17, Long Beach, California (2017)

\bibitem{DBLP:journals/corr/VinyalsL15}
Vinyals, O., Le, Q.V.: A neural conversational model. CoRR
  \textbf{abs/1506.05869} (2015)

\end{thebibliography}

\end{document}